# Deep-Sea A*+: An Advanced Path Planning Method Integrating Enhanced A* and Dynamic Window Approach for Autonomous Underwater Vehicles

Yinyi Lai[1], Jiaqi Shang[1], Zenghui Liu[1,*], Zheyu Jiang[1], Yuyang Li[2], and Longchao Chen[2]

[1] Department of Mechanical Engineering, Hohai University, Nanjing 211100, China
[2] Intelligent Manufacturing Engineering, Hohai University, Changzhou 213022, China
*Corresponding author: Zenghui Liu (E-mail: zenghui.liu@hhu.edu.cn)

*Abstract*—As terrestrial resources become increasingly depleted, the demand for deep-sea resource exploration has intensified. However, the extreme conditions in the deep-sea environment pose significant challenges for underwater operations, necessitating the development of robust detection robots. In this paper, we propose an advanced path planning methodology that integrates an improved A* algorithm with the Dynamic Window Approach (DWA). By optimizing the search direction of the traditional A* algorithm and introducing an enhanced evaluation function, our improved A* algorithm accelerates path searching and reduces computational load. Additionally, the path-smoothing process has been refined to improve continuity and smoothness, minimizing sharp turns. This method also integrates global path planning with local dynamic obstacle avoidance via DWA, improving the real-time response of underwater robots in dynamic environments. Simulation results demonstrate that our proposed method surpasses the traditional A* algorithm in terms of path smoothness, obstacle avoidance, and real-time performance. The robustness of this approach in complex environments with both static and dynamic obstacles highlights its potential in autonomous underwater vehicle (AUV) navigation and obstacle avoidance.

## I. INTRODUCTION

The depletion of terrestrial resources has shifted human exploration towards the deep sea, which holds immense untapped resource potential. However, the deep-sea environment presents formidable challenges, including high pressure, low temperatures, darkness, and strong corrosiveness. These extreme conditions make traditional manual operations not only inefficient but also hazardous [1]. As a result, developing robots capable of operating in these harsh environments is of paramount importance [2]. Such robots are widely employed in applications like marine exploration, environmental monitoring, seabed mining, and pipeline inspections [3], [4], [5].

Path planning for underwater robots introduces several unique challenges. The seabed environment is highly complex, involving both static obstacles, such as underwater mountains and reefs, and dynamic ones such as fish schools and marine life. This necessitates robust dynamic obstacle avoidance capabilities. Additionally, underwater environments are inherently uncertain, with fluctuating water currents that constrain a robot's ability to execute sharp turns, thus demanding smooth path planning. Energy efficiency is also a critical consideration for deep-sea missions, as underwater robots must operate for extended periods with limited energy reserves [6]. Furthermore, the lack of GPS signals underwater forces these robots to rely on sensors such as sonar and lidar for navigation, complicating the perception system.

Given these challenges, the autonomy and intelligence of underwater robots are directly linked to the success of their missions. Robots must autonomously navigate and avoid obstacles while ensuring that the planned path is smooth, efficient, and safe [7]. Achieving this requires the path planning algorithm to not only perform global planning but also handle real-time obstacle avoidance in dynamic environments. Particularly in deep-sea scenarios, the precision, smoothness, and real-time nature of the path are crucial to the successful navigation of AUVs [8].

Robot path planning has been extensively researched [9], [10], [11], with current methods typically categorized into two main approaches: global and local path planning. Global path planning methods are applied in known environments to compute the shortest path between a starting point and a target [12]. In contrast, local path planning focuses on real-time obstacle avoidance, leveraging sensor data to dynamically adapt to environmental changes [13].

Among global path planning methods, well-established algorithms such as A* algorithm [14], D* algorithm [15],

Rapidly-exploring Random Tree (RRT) [16], Dijkstra algorithm [17], and Ant Colony Optimization algorithm [18], are widely utilized in robotic navigation systems. Dijkstra's algorithm ensures the optimal path by traversing the smallest child node through the principle of greed, but its traversal of all node information requires a significant amount of time and has low operational efficiency [19], making it unsuitable for underwater environments. X. Wu et al. proposed a bidirectional A* algorithm that can expand from both the starting point and the target point simultaneously, enhancing the algorithm's convergence speed, but it does not account for the interference of dynamic obstacles [20]. R. Son et al. employed three Smoothers to reduce the number of turning points and eliminate redundant path nodes, but this method is susceptible to the influence of node quantity and necessitates multiple iterations [21].

On the local planning front, it aims to calculate the robot's next movement trajectory by perceiving environmental changes in real time. Common local path planning algorithms include the Dynamic Window Approach (DWA) [22], Artificial Potential Field (APF) [23], and Model Predictive Control (MPC) [24]. These algorithms excel in real-time performance and adaptability to dynamic environments. However, DWA, which only considers local path planning, has poor global navigation capabilities and is prone to getting trapped in local optima in complex environments. The APF method uses potential functions across the entire configuration space to guide the robot's movement, but it typically ends in local minima [25]. Shen. Y et al. introduced a Q-learning reinforcement learning algorithm for selecting appropriate DWA target function weights in different environments, effectively enhancing the applicability of DWA in complex environments [26].

Therefore, integrating global and local path-planning algorithms to address path-planning challenges in complex undersea environments has become a focal point in current underwater robot research [27].

This paper introduces a path planning method that combines an improved A* algorithm with the DWA, aiming to address the autonomous navigation challenges of deep-sea underwater robots in complex marine environments. The specific contributions of this paper are as follows:

(1) Enhancement of the traditional A* algorithm: The search path directions are optimized from eight to five, a new evaluation function is employed, and path smoothness is improved, which successfully reduces computational load, accelerates path search speed, and thus enhances the responsiveness of underwater robots. Additionally, it ensures that underwater robots will not diagonally traverse the vertices of obstacles during path planning, effectively mitigating potential collision risks.

(2) Integration of global and local planning: We propose an improved A* algorithm that combines global path planning with local dynamic obstacle avoidance methods DWA. Our approach has been validated through complex simulation experiments, demonstrating its ability to achieve autonomous, efficient, and safe underwater robot missions in complex dynamic environments.

II. METHODS

In this paper, we propose a path-planning method for deep-sea underwater vehicles that combines an improved A* algorithm with the DWA. The method balances path smoothness, obstacle avoidance, and real-time performance by utilizing a two-stage planning process, wherein global and local path planning are seamlessly integrated.

A. *Improved A* Algorithm*

The A* algorithm is a well-established heuristic search method that utilizes a cost function to estimate the distance from the current node to the goal node, guiding the search process accordingly. While it ensures the discovery of an optimal path in static environments, it also enhances efficiency by relying on heuristic evaluation, making it suitable for scenarios requiring precise planning. In this study, the A* algorithm is employed to compute the global path. To address the specific challenges faced by deep-sea underwater vehicles, several enhancements to the traditional A* algorithm are proposed:

*1) Heuristic Function Optimization:* The heuristic function of the A* algorithm is designed based on the Euclidean distance, ensuring that the heuristic value never exceeds the actual distance from the current point to the goal point. Traditional A* algorithm searches through many unnecessary nodes during path planning, affecting search efficiency. This implies that when the heuristic function's estimated value is greater than the actual value, fewer search nodes are required, resulting in high efficiency, but it may not find the optimal path. Conversely, when the heuristic function's estimated value is less than the actual value, it can find the optimal path but requires searching through more nodes, leading to lower computational efficiency. The highest search efficiency is achieved only when the heuristic function's estimated value equals the actual value. Therefore, we introduce a new evaluation function:

$$f(n) = g(n) + (1 - \log(P)) \cdot h(n) \quad (1)$$

where $f(n)$ represents the comprehensive cost value; $g(n)$ is the actual cost from the starting point to the current node; $h(n)$ is the estimated cost from the current node to the goal. $P$ represents the obstacle rate between the starting point and the goal point, calculated as the number of obstacles divided by the total number of grids.

*2) Search Point Selection Strategy:* The traditional A* algorithm evaluates the eight neighboring grid cells of the current node, as shown in Figure 1, where each of the eight directions (1-8) represents a potential move. However, many of these directions are redundant when the robot's relative position to the goal is considered. To improve computational efficiency, our approach restricts the expansion to five directions based on the goal's orientation relative to the current node. This significantly reduces computational overhead by eliminating unnecessary search points.

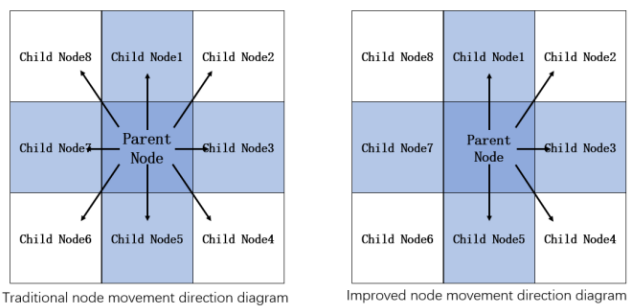

Figure 1. Schematic Diagrams of Search Path Directions

*3) Path Smoothing:* The paths generated by traditional A* often include abrupt turns, increasing the complexity of execution for real-time systems. To address this, we apply the Floyd algorithm to smooth the generated paths. This algorithm identifies and eliminates nodes that contribute minimally to overall path smoothness, reducing the number of sharp turns and improving continuity. The resulting path is both more efficient and easier for the robot to follow, ultimately enhancing the robot's stability and reducing energy consumption during operation.

B. *Dynamic Window Approach*

The DWA is a local obstacle avoidance algorithm based on the robot's dynamic constraints, capable of performing real-time obstacle avoidance during the robot's motion. The core idea of the DWA algorithm is to search the robot's possible velocity space within a given time window to find the optimal velocity combination that can safely avoid obstacles and head toward the goal point. This paper combines the DWA with the improved A* algorithm, enabling the robot to dynamically respond to encountered obstacles during the execution of the global path.

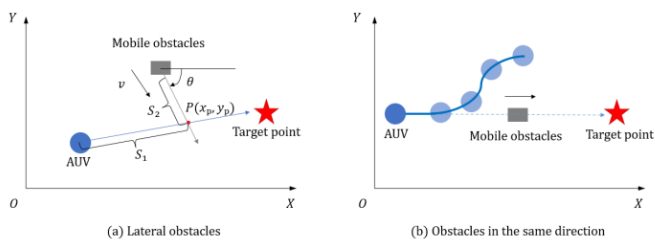

Figure 2. Schematic Diagrams of AUV and Mobile Obstacle Movement

The DWA algorithm ensures that the generated path is within the dynamic constraints by imposing limits on the robot's speed and acceleration. Specifically, the DWA algorithm generates a range of possible combinations of speed and acceleration and evaluates each velocity combination. The evaluation criteria include collision risk, distance to the goal, and the smoothness of the trajectory. In this way, DWA can adjust the robot's motion trajectory in real time to avoid collisions with dynamic obstacles while ensuring safety. As shown in Figure 2, P represents a virtual collision point. Based on the virtual collision point P, we calculate the time $t_1$ it takes for the AUV and the obstacle to reach point P, and also determine whether the obstacle is outside the safe distance upon arrival.

Depending on the movement directions of the robot and the obstacle, we categorize dynamic obstacle avoidance path planning into two types. The first type is lateral obstacle avoidance, as shown in Figure 3; the second type is head-on obstacle avoidance, as shown in Figure 4. Based on the movement intentions of the obstacles, the robot makes a judgment on whether a collision will occur.

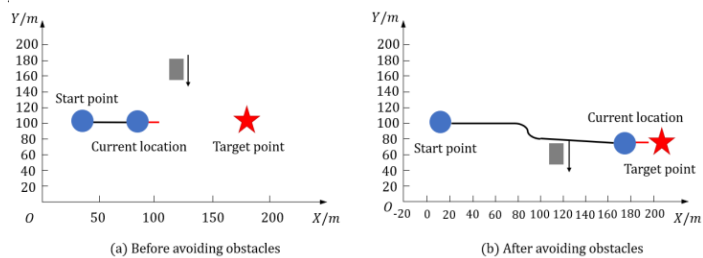

Figure 3. Schematic Diagrams of Lateral Obstacle Avoidance Algorithm

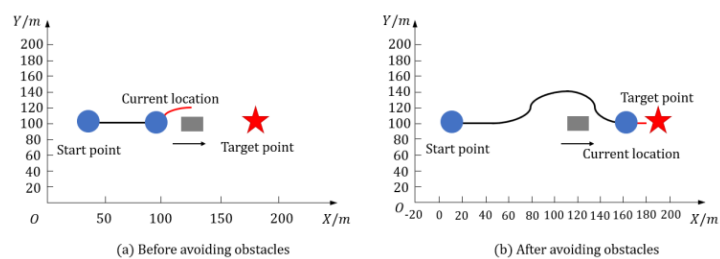

Figure 4. Schematic Diagrams of Head-on Obstacle Avoidance Algorithm

III. SIMULATION EXPERIMENTS AND RESULTS

To validate the effectiveness of the proposed method, a series of simulation experiments were conducted in Matlab 2023a, emulating a complex deep-sea environment with both static and dynamic obstacles. The underwater robot's task was to navigate from a starting point to a designated target while avoiding all obstacles. The performance of the traditional A* algorithm was compared with the improved A* algorithm integrated with DWA across several key metrics, including path smoothness, obstacle avoidance, and real-time adaptability.

A. *Experimental Setup*

The experiments were conducted in two grid environments of different sizes: 30x30 and 20x20. In each grid, we simulated the path planning for an AUV from the starting point to the target point. During the path planning process, the AUV needed to avoid both dynamic and static obstacles. We compared the performance of the traditional A* algorithm and the improved A* algorithm in terms of the number of traversed nodes, path length, number of turns, and turn angles.

## B. Scene without Moving Obstacles

TABLE I. PATH PLANNING RESULTS WITHOUT MOVING OBSTACLES (20*20)

| 20 * 20 grid | A* | Improved A* |
|---|---|---|
| Traverse the number of nodes | 85.00 | 55.00 |
| Algorithm path length (m) | 19.56 | 19.89 |
| Number of turning points | 6.00 | 4.00 |
| Turning angle (rad) | 270.00 | 167.04 |

TABLE II. PATH PLANNING RESULTS WITHOUT MOVING OBSTACLES (30*30)

| 30 * 30 grid | A* | Improved A* |
|---|---|---|
| Traverse the number of nodes | 256.00 | 160.00 |
| Algorithm path length (m) | 40.28 | 40.16 |
| Number of turning points | 11.00 | 5.00 |
| Turning angle (rad) | 495.00 | 159.34 |

To assess the performance of the proposed method in obstacle-free environments, we conducted short path planning experiments on a 20x20 grid and long path planning experiments on a 30x30 grid. Both the improved A* algorithm and the traditional A* algorithm were capable of effectively planning the shortest path from the starting point to the target point.

However, as shown in Tables 1 and 2, the improved A* algorithm demonstrated superior performance in terms of the number of nodes traversed and planning time. Specifically, in the 20x20 grid, the improved A* algorithm traversed 55 nodes, compared to 85 for the traditional A* algorithm, representing a 35.3% reduction. In the 30x30 grid, the improved A* algorithm traversed 160 nodes, compared to 256 for the traditional A* algorithm, a 37.5% reduction. This indicates that the improved A* algorithm can plan effective paths with fewer computational resources in obstacle-free scenarios.

Additionally, as evident from Figure 5(b) and 6(b), the yellow path (traditional A* algorithm) is more likely to collide with the vertices of obstacles compared to the purple path (improved A* algorithm). This further demonstrates the superiority of our method.

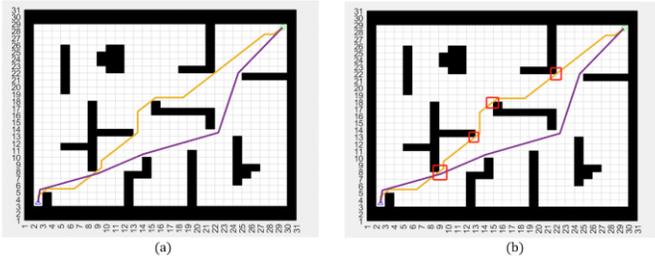

Figure 5. 30 * 30 Grid Path Planning Algorithm Diagrams

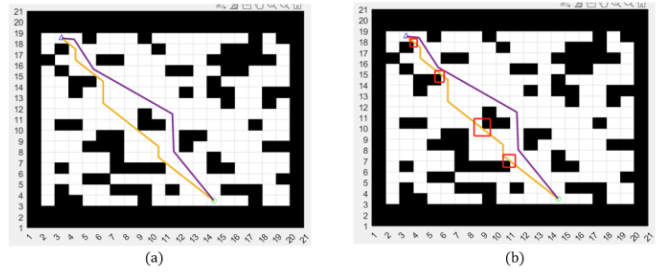

Figure 6. 20 * 20 Grid Path Planning Algorithm Diagrams

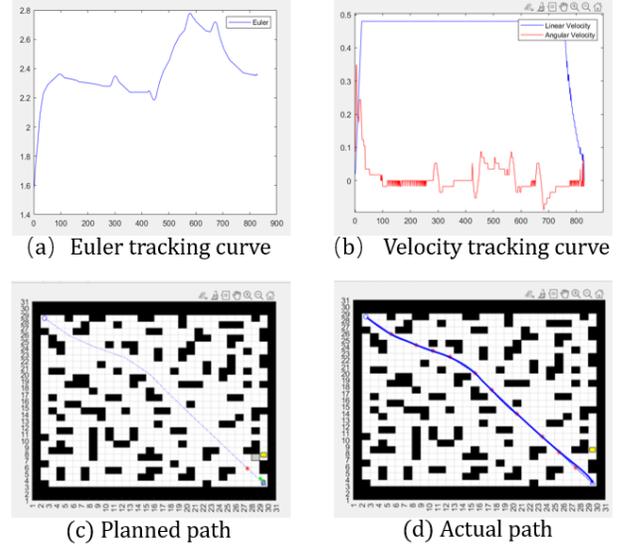

(a) Euler tracking curve  (b) Velocity tracking curve

(c) Planned path  (d) Actual path

Figure 7. Results of Planned and Actual Paths without Moving Obstacles

We also conducted simulation experiments incorporating the DWA-integrated improved A* algorithm in scenarios where moving obstacles do not interfere with the path. The results are shown in Figure 7 and Table 3. It can be observed that the actual operating path of the AUV in the simulation experiment almost completely overlaps with the planned path, with minor initial deviations due to the inevitable directional adjustment errors caused by the initial direction setting of the underwater robot being inconsistent with the target direction.

TABLE III. ALGORITHM RESULTS WITHOUT INTERFERENCE FROM MOVING OBSTACLES

| 30 * 30 grid | Improved A * algorithm integrating DWA |
|---|---|
| Total time (s) | 549.49 |
| Total distance traveled (m) | 37.39 |
| Traverse the number of nodes | 112.00 |
| Number of turning points | 4.00 |
| Turning angle (rad) | 85.58 |
| Planning time (s) | 0.07 |

TABLE IV. IMPROVED A * ALGORITHM INTEGRATING DWA RESULTS WITH INTERFERENCE FROM DIFFERENT MOVING OBSTACLES

| 30 * 30 grid | lateral obstacle | Head-on Obstacle |
|---|---|---|
| Total time (s) | 558.27 | 556.35 |
| Total distance traveled (m) | 38.35 | 37.70 |

*C. Lateral Obstacle Scenario*

In this scenario, the AUV faces dynamic obstacles that appear laterally to its initial path. To further increase the complexity of the environment, we introduced static obstacles that the AUV must avoid while still tracking its overall planned trajectory. The lateral obstacle avoidance test aims to evaluate the real-time adaptability of the improved A* algorithm integrated with the DWA when encountering obstacles on the side of the AUV's trajectory.

As depicted in Figure 8, the AUV begins by detecting a lateral obstacle shortly after starting its path. The DWA algorithm computes a range of possible velocity and acceleration combinations within a short time window, dynamically adjusting the AUV's course to avoid the obstacle. The AUV smoothly bypasses the lateral obstacle without making any abrupt changes to its path. As the robot approaches the newly introduced static obstacle, the algorithm again recalculates the optimal path. By avoiding this static obstacle, the AUV successfully returns to the initial planned path, proceeding towards its target. The attitude angle and velocity tracking curves during the experiment are shown in Figure 8(g) and 8(h).

The detailed Path Analysis is as follows:

*1) Collision Avoidance:* During the lateral obstacle avoidance phase, the algorithm's ability to adjust the AUV's velocity and acceleration ensured that the robot maintained a safe distance from the obstacle without deviating significantly from the optimal path.

*2) Smoothness of Motion:* One of the most critical features observed in this scenario was the smoothness with which the AUV maneuvered around obstacles. The path-smoothing capabilities of the improved A* algorithm, combined with the real-time adaptability of DWA, resulted in minimal sharp turns, enhancing the continuity of the AUV's movement.

*3) Re-alignment with Planned Path:* After bypassing both dynamic and static obstacles, the AUV promptly re-aligned with the pre-planned global path, maintaining a direct trajectory to the target. This demonstrates the algorithm's robust capability to recalibrate without excessive computational delays.

*4) Simulation Results:* In the 30x30 grid environment, the AUV completed the task in 558.27 seconds, covering a total distance of 38.35 meters. Notably, the improved A* algorithm with DWA enabled the robot to avoid the lateral obstacle while keeping the number of turning points to a minimum (4 turning points) and reducing the total turn angle to 85.58 degrees. This performance highlights the method's efficiency in minimizing unnecessary detours, ensuring that the AUV maintains optimal path smoothness even in the presence of complex, multi-obstacle environments.

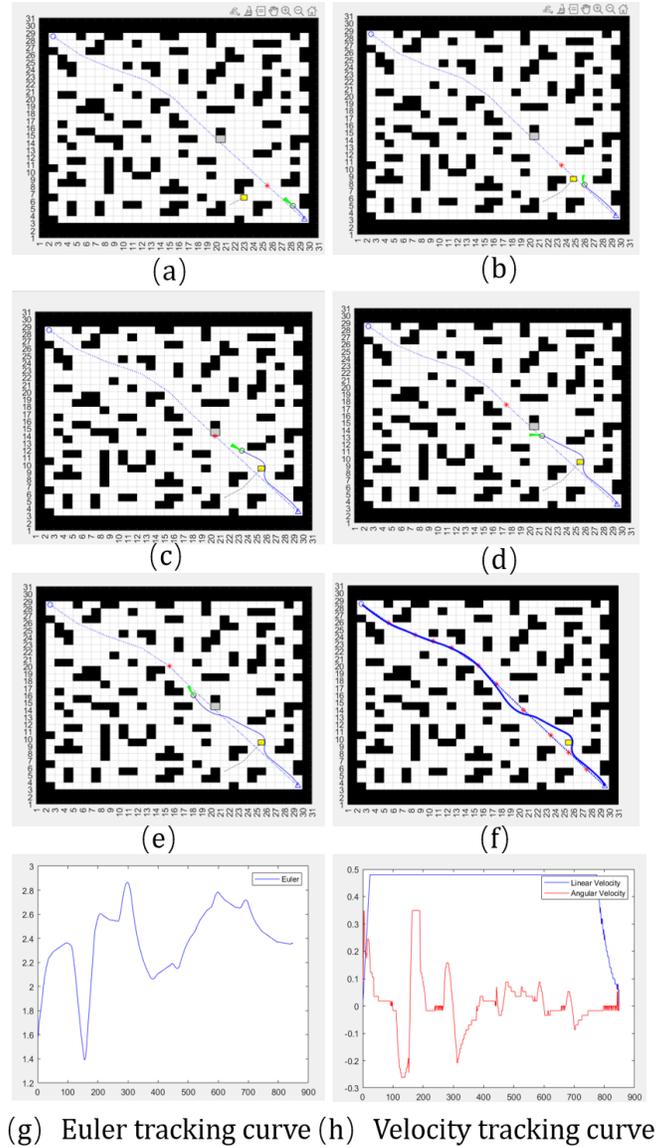

(a) (b) (c) (d) (e) (f)

(g) Euler tracking curve (h) Velocity tracking curve

Figure 8. Results for Lateral Obstacle

*D. Head-on Obstacle Scenario*

In the head-on obstacle avoidance scenario, the AUV faces a dynamic obstacle moving in the same direction along its intended path, creating a higher risk of collision. This setup simulates real-world situations where underwater vehicles encounter other moving objects, such as marine life or other robotic vehicles, that require more sophisticated planning to ensure safe passage.

However, the improved A* algorithm combined with the DWA algorithm still showed good performance. The head-on scenario adds complexity as the AUV must not only avoid a direct collision but also determine the most efficient strategy to bypass the obstacle, either by slowing down, overtaking, or altering its path entirely. From Figure 9(b), we can see the AUV begins to avoid the head-on obstacle and successfully aligns with it by Figure 9(c). Subsequently, as shown in Figure 9(d), the AUV overtakes the head-on obstacle and is about to return to the initial planned route. Finally, as depicted in Figure 9(e), it completes the avoidance of the new static obstacle and reaches the destination. The initial and actual routes of the AUV are shown in Figure 9(f). Our experiment demonstrates that the improved A* algorithm combined with DWA can effectively plan a path that avoids head-on obstacles. In the 30x30 grid, the total time for the improved A * algorithm integrating DWA was 558.27 seconds, with a total distance of 38.35 meters. The attitude angle and velocity tracking curves during the experiment are shown in Figure 9(g) and 9(h).

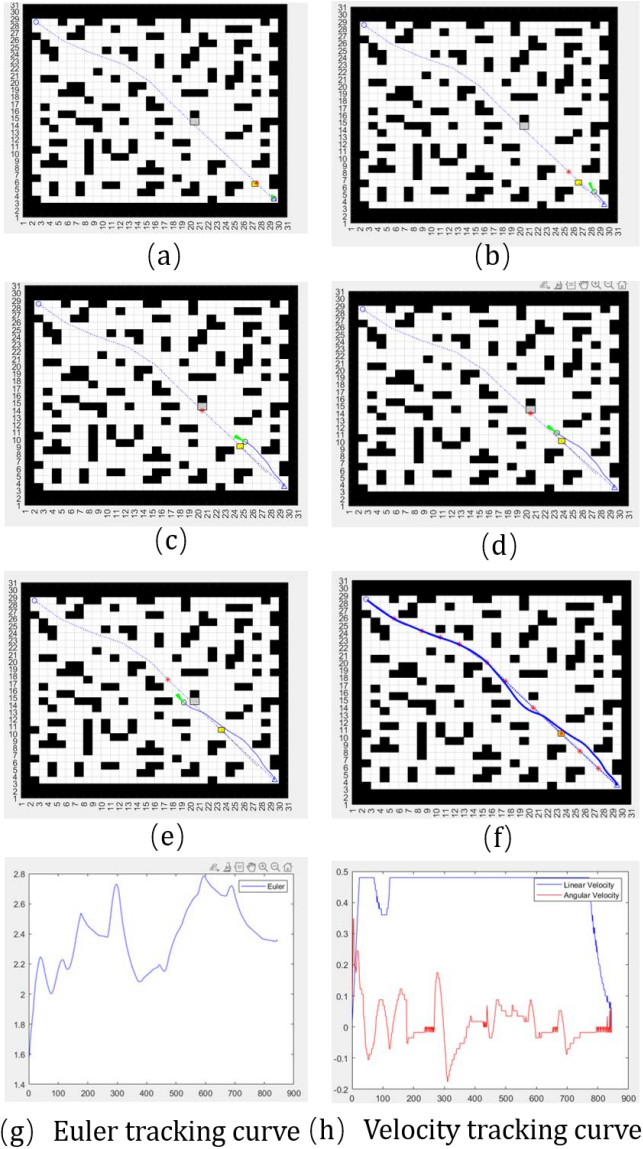

(a)　(b)　(c)　(d)　(e)　(f)

(g) Euler tracking curve　(h) Velocity tracking curve

Figure 9. Results for Head-on Obstacle

In summary, our improved A* algorithm integrated with DWA has shown superior performance in various scenarios, whether in simple obstacle-free environments or complex environments with dynamic and static obstacles. It can effectively plan the optimal path and maintain high precision in actual path tracking. These results prove the potential application of the improved A* algorithm in the field of path planning, especially in the navigation and obstacle avoidance of mobile robots such as AUVs.

IV. CONCLUSION AND FUTURE WORK

This paper presents a path planning method for deep-sea underwater robots that combines an improved A* algorithm with DWA, effectively addressing path planning and dynamic obstacle avoidance in complex marine environments. Simulation experiments have verified the method's superiority in terms of path smoothness, obstacle avoidance capability, and real-time performance, making it suitable for deep-sea underwater robot missions.

Our future research will explore the integration of additional sensor technologies and neural network algorithms that are currently applied in underwater imaging [28] and AUV path planning [29], [30], to further enhance the precision and reliability of path planning.